\documentclass{article}

\usepackage{PRIMEarxiv}

\usepackage[utf8]{inputenc} % allow utf-8 input
\usepackage[T1]{fontenc}    % use 8-bit T1 fonts
\usepackage{hyperref}       % hyperlinks
\usepackage{url}            % simple URL typesetting
\usepackage{booktabs}       % professional-quality tables
\usepackage{amsfonts}       % blackboard math symbols
\usepackage{nicefrac}       % compact symbols for 1/2, etc.
\usepackage{microtype}      % microtypography
\usepackage{lipsum}
\usepackage{fancyhdr}       % header
\usepackage{graphicx}       % graphics
\graphicspath{{media/}}     % organize your images and other figures under media/ folder
\usepackage{amsmath}
\usepackage{subcaption}

\title{Ethereum Price Prediction Employing Large Language Models for Short-term and Few-shot Forecasting
}

\author{
\textbf{Eftychia Makri}$^{1}$, 
\textbf{Georgios Palaiokrassas}$^{1}$, 
\textbf{Sarah Bouraga}$^{2}$, 
\textbf{Antigoni Polychroniadou}$^{3,4}$, 
\textbf{Leandros Tassiulas}$^{1}$ 
\\
\\
$^{1}$Department of Electrical Engineering, Yale University, New Haven, CT, USA \\ 
$^{2}$Metis Lab, EM Normandie Business School, Paris, France \\ 
$^{3}$J.P. Morgan AI Research, New York, NY, USA \\ 
$^{4}$J.P. Morgan AlgoCRYPT CoE, New York, NY, USA \\ 
\\
\texttt{\{eftychia.makri, george.palaiokrassas, leandros.tassiulas\}@yale.edu} \\ 
\texttt{sbouraga@em-normandie.fr} \\ 
\texttt{antigoni.polychroniadou@jpmorgan.com} 
}

\begin{document}
\maketitle

\begin{abstract}
Cryptocurrencies have transformed financial markets with their innovative blockchain technology and volatile price movements, presenting both challenges and opportunities for predictive analytics. Ethereum, being one of the leading cryptocurrencies, has experienced significant market fluctuations, making its price prediction an attractive yet complex problem. This paper presents a comprehensive study on the effectiveness of Large Language Models (LLMs) in predicting Ethereum prices for short-term and few-shot forecasting scenarios. The main challenge in training models for time series analysis is the lack of data. We address this by leveraging a novel approach that adapts existing pre-trained LLMs on natural language or images from billions of tokens to the unique characteristics of Ethereum price time series data. Through thorough experimentation and comparison with traditional and contemporary models, our results demonstrate that selectively freezing certain layers of pre-trained LLMs achieves state-of-the-art performance in this domain. This approach consistently surpasses benchmarks across multiple metrics, including Mean Squared Error (MSE), Mean Absolute Error (MAE), and Root Mean Squared Error (RMSE), demonstrating its effectiveness and robustness. Our research not only contributes to the existing body of knowledge on LLMs but also provides practical insights in the cryptocurrency prediction domain. The adaptability of pre-trained LLMs to handle the nature of Ethereum prices suggests a promising direction for future research, potentially including the integration of sentiment analysis to further refine forecasting accuracy.
\end{abstract}

% keywords can be removed

\section{Introduction}
\label{sec:Intro}
Cryptocurrencies have been part of our reality for more than a decade now. We have witnessed a rapid evolution of the cryptocurrency market in a short period of time and it has become a stimulating research area. One of the reasons cryptocurrencies became popular is their ability to mitigate some limitations of the traditional fiat currencies, such as its dependence on central institutions and issues pertaining to trust, security, transparency, and flexibility \cite{patel2022fusion}. 

The global cryptocurrency market is valued today at \$2.49 Trillion \footnote{\url{https://www.forbes.com/digital-assets/crypto-prices/}} and has slowly pushed towards its maturity since 2014 \cite{wkatorek2021multiscale}. It attracts many investors, whose behavior is influenced by social influence or public sentiment \cite{almeida2023systematic}. Cryptocurrency trading strategy has been extensively studied \cite{fang2022cryptocurrency}. Market booms, crisis periods and media attention have led to research analyzing the dynamics of stress periods, return characteristics and cryptocurrency market efficiency (or lack thereof). A study shows that, when considering the length of the efficiency period of cryptocurrency markets, Bitcoin's is generally more efficient than Ethereum's \cite{noda2021evolution,mokni2024efficiency}. 

One prominent characteristics of cryptocurrencies is their high volatility. Case in point, Bitcoin was valued at less than \$40.000 in January 2024 to reach almost \$67.000 three months later in March 2024. Investors perceive this high volatility positively, i.e. as a chance to reach higher profits \cite{almeida2023systematic}. 
This volatility characteristic has also sparked the interest of researchers, leading to numerous studies proposing approaches for price prediction. Existing research use market data (including information such as price, trading volume, or order-level) and/or sentiment-based data (such as social media data) \cite{fang2022cryptocurrency}. Because of the lack of seasonal effect in cryptocurrency, traditional statistical approaches have been proven inefficient \cite{khedr2021cryptocurrency}. Hence, many scholars have explored the use of Machine Learning (ML), Deep Learning (DL) or reinforcement and federated learning for price prediction \cite{patel2022fusion}, with a focus primarily on Bitcoin \cite{khedr2021cryptocurrency}. From existing studies, it appears that Artificial Neural Networks (ANN) \cite{akyildirim2021prediction}, \cite{yiying2019cryptocurrency} and Random Forest (RF) \cite{khedr2021cryptocurrency} have shown promising results as ML classifiers for cryptocurrency price prediction, while Support Vector Machine (SVM) has been particularly effective for Ethereum \cite{poongodi2020prediction}. As far as DL is concerned, Long Short-Term Memory (LSTM) was one of the most appropriate approaches to address the limitations observed with Multilayer Perceptron (MLP) and Recurrent Neural Network (RNN), namely the difficulty for MLP to recognize the temporal element of time series, and the vanishing gradient problem occurring with both MLP and RNN \cite{jay2020stochastic}. LSTM was thus established as one of the most suitable techniques for time-series prediction problems \cite{khedr2021cryptocurrency}. 

A recent advancement in Artificial Intelligence (AI) that has gained tremendous attention, both in the research community and in the industry, is Large Language Models (LLMs). LLMs have been applied for various tasks in research, and also in various domains, such as finance \cite{zhao2023survey}. Lately, those models are being employed for a direct processing of timeseries data \cite{jiang2024empowering}, \cite{tan2024language}, \cite{li2024foundts}. In this direction, research efforts such as Time-LLM \cite{jin2023timellm}, LLM4TS \cite{chang2023llm4ts} and GPT2(6) \cite{zhou2023one}, adopt a method where they freeze the LLM encoder backbones and simultaneously fine-tune or adapt the timeseries input and distribution heads specifically for forecasting tasks. In this paper, we build upon these approaches by exploring the use of pre-trained LLMs, specifically Llama-3, Llama-2 and GPT-2, by applying the technique of freezing certain layers while fine-tuning others for the challenging task of Ethereum price prediction. Hence, the goals and corresponding \textbf{contributions} of this paper are the following: 
\begin{itemize}
    \item \textbf{We show that LLMs}, could work comparably or even better than state-of-the-art (SOTA) models for cryptocurrency prediction.
    \item \textbf{We test and expand the universality of \cite{zhou2023one}} by exploring its power as an Ethereum price forecaster, with various LLMs as backbone models.
    \item \textbf{We are the first to explore} the capabilities of Llama and GPT for directly processing Ethereum time series data, rather than using sentiment analysis.
    \item \textbf{We explore the few-shot learning abilities} of LLMs in a context of cryptocurrency price prediction.
\end{itemize}
In order to address our research questions, we adapt and extend the framework outlined in \cite{zhou2023one}. We collect time series Ether price data, we test various models and forecasting methods, and compare our results to the SOTA. The main findings of this study show that LLMs, when appropriately fine-tuned, can compete with traditional SOTA models. 

With this study, we contribute to the body of knowledge in Decentralized Finance (DeFi) by exploring the use of LLMs for cryptocurrency price prediction. While most studies examined the task of Bitcoin price prediction, our contribution lies in the exploration of Ether price prediction. The results can lead to a toolkit for practitioners interested in price prediction, including financial analysts, investors, and traders. 

The remainder of this paper is structured as follows. Section \ref{related} presents the related work. Section  \ref{sec:method} introduces the methodology while Section \ref{sec:evaluation} describes the evaluation criteria. Section \ref{sec:results} presents our results. Finally Section \ref{sec:conclusion} discusses the results and concludes this paper.
\section{Related Work}
\label{related}

Cryptocurrencies have undergone substantial fluctuations in price, which complicates the task of predicting their future values but also makes accurate predictions highly valuable for trading. Many research efforts have investigated the application of ML algorithms, DL frameworks, and sentiment analysis in predicting the prices of cryptocurrencies. Nonetheless, existing studies often face limitations and overlook certain aspects, failing to take into account a variety of factors or to apply more sophisticated methods.

Oyedele et al. \cite{oyedele2023performance} conducted a performance evaluation of DL, including Convolutional Neural Networks (CNNs) and Gated Recurrent Units (GRUs), and boosted tree-based techniques for predicting cryptocurrency closing prices. They focused on the optimization of these models using a genetic algorithm to enhance prediction accuracy on datasets from multiple cryptocurrencies. Akyildirim et al. \cite{akyildirim2021prediction} explored the predictability of the most liquid twelve cryptocurrencies at daily and minute-level frequencies by employing SVM, Logistic Regression (LR), ANNs, and RF. The study demonstrated that using past price information and technical indicators as model features could achieve predictive accuracy consistently above the 50\% threshold, indicating a certain degree of predictability in cryptocurrency market trends. Both studies did not use the most established metrics for time series, such as mean squared error (MSE), root mean squared error (RMSE), and mean absolute error (MAE), making it difficult to compare their results with other works.

Kumar et al. \cite{kumar2020predicting} employed MLP and LSTM to predict the daily prices of Ethereum. The data used were collected on a daily basis from August 2015 to August 2018, while the targeted value was the "Open" price of the day. In the same direction, Kim et al.\cite{kim2021predicting} used ANNs to identify significant variables influencing Ethereum prices, from August 2015 to November of 2018, and to demonstrate the effectiveness of ANNs in capturing and predicting price fluctuations. Our work, outperforms those studies in terms of MSE and RMSE, highlighting the importance of our method.

\subsection{Cryptocurrency Price Prediction with Transformers}
In \cite{sridhar2021multi}, the authors utilize a transformer model with multi-head attention in the encoder-decoder architecture to forecast Dogecoin prices on an hourly basis. The model's accuracy is assessed using various metrics, such as MAE and the predictive R-squared value. It is compared with state-of-the-art models for cryptocurrency prediction, thus their study is limited to the Dogecoin prices. The approach in \cite{tanwar2022prediction} employs Transformer and LSTM networks to predict the prices of different cryptocurrencies. While incorporating LSTM with Transformers results in increased computational time, the accuracy of predictions is improved compared to conventional regression neural networks and k-Nearest Neighbors (k-NN) forecasting models. Although this is a very interesting work, it only uses 180 samples, corresponding to days of Ethereum price fluncuations, leading to a comparably high RMSE.

Son et al. \cite{son2022using} employ Natural Language Processing (NLP) to conduct stance detection regarding specific entities for prediction purposes. The next phase involves applying the detected stances to actual price data, utilizing an RNN to convert stance information into price forecasts. The stance detection system, powered by RoBERTa, attained an 80\% accuracy rate. Separately, a price forecasting model based on RNNs reported a mean absolute error of \$1,144, which is relatively minor given that cryptocurrency prices can soar to \$60,000. While transformer models are used in this work, only the RNN model is used for a direct process of the timeseries data.

Singh et al. \cite{singh2024transformer} explored the effectiveness of a transformer-based neural network in predicting Ethereum cryptocurrency prices. Their work was based on the hypothesis that the prices of cryptocurrencies have a strong correlation with each other and are influenced by the surrounding sentiment. The model utilized a transformer framework in various setups, ranging from simple single-feature cases to more intricate designs that include trading volume, sentiment analysis, and the prices of correlated cryptocurrencies, showing that by predicting only the Ethereum prices and by not taking into account the sentiment analysis scores, can lead to an improved performance.

\subsection{LLMs as Timeseries Forecasters}

Xie et al. \cite{ZeroShotAnalysisChatGPTStockPrediction} conducted a zero-shot analysis of ChatGPT's performance in predicting stock market movements based on tweets and historical stock price datasets. This study demonstrated that ChatGPT, despite its strong language understanding abilities, underperformed in this financial domain compared to SOTA methods and even basic linear regression models. This research underscored the challenges and potential limitations of applying ChatGPT, specifically in the context of financial market prediction tasks, highlighting the need for more specialized training or fine-tuning to enhance its applicability in such complex domains.

Rasul et al. \cite{rasul2023lag} introduced Lag-Llama, a model for univariate probabilistic time series forecasting. It was designed to investigate the scaling behavior of time series foundation models, showing impressive zero-shot prediction capabilities on unseen datasets, outperforming supervised baselines in certain conditions. This model was distinct for its use of lag-features, diverging from other pre-trained models by focusing on univariate forecasting rather than multivariate, which is common in the field.

In the same direction, it was demonstrated how GPT-3 and LLaMA-2 can effectively perform zero-shot time series forecasting by encoding time series as strings of numerical digits, treating forecasting as next-token prediction \cite{gruver2023large}. This innovative approach, dubbed LLMTIME, challenged traditional time series models by leveraging the inherent capabilities of LLMs without requiring task-specific training or fine-tuning.

A novel framework for enhancing time-series forecasting using pre-trained LLMs like GPT-2 was proposed in \cite{chang2023llm4ts}. It focused on aligning LLMs to the specifics of time-series data through a two-stage fine-tuning strategy, incorporating a novel two-level aggregation method to integrate multi-scale temporal data effectively. This approach enabled LLMs to better interpret time-specific information, leading to superior forecasting performance in both full and few-shot scenarios compared to existing methods.

The paper in \cite{pan2024s} introduces a novel approach to enhance time series forecasting by aligning time series embeddings with the semantic space of pre-trained LLMs. It proposes a tokenization module that decomposes time series data into patches, capturing temporal dynamics effectively and enabling their alignment with the semantic space of LLMs. By leveraging pre-trained word token embeddings as semantic anchors, the model maximizes cosine similarity in the joint space, allowing it to retrieve relevant semantic prompts that guide forecasting decisions. The proposed $S^2$IP-LLM model outperforms state-of-the-art baselines across multiple benchmark datasets, with ablation studies and visualizations confirming the benefits of integrating semantic space-informed prompt learning.

Zhou et. al \cite{zhou2023one}, explored the utilization of pre-trained models, without modifying the core architecture, for time series analysis tasks. A notable approach is the Frozen Pretrained Transformer (FPT), which retains the original configuration of self-attention and feedforward layers from pre-trained language or image models. This method involves fine-tuning these comprehensive models, in this case GPT-2, on a wide array of time series tasks. Findings from such studies indicate that leveraging pre-trained models, originally developed for processing natural language or images, can produce performance on par with or superior to current leading methods in various time series analysis challenges.

In this paper, we draw on this framework for the application of LLMs on Ether price prediction. Furthermore, we expand 
\cite{zhou2023one} work, by experimenting not only with GPT-2 as backbone model, but with Llama-2 as well. To the best of our knowledge, this is the first time LLMs are employed for a direct process of timeseries data for a cryptocurrency price prediction task.

\section{Method} \label{sec:method}
Inspired by the success of \cite{zhou2023one} in timeseries forecasting as described in Section \ref{related}, we adopt their framework to explore the use of LLMs, like Llama and GPT-2, for predicting Ethereum prices.
\subsection{Dataset}
Data used for this research come from publicly available Ethereum price datasets as well as Ethereum transactions we retrieved. To this direction, we have set up an Ethereum full archive node (8-core Intel i7-11700 CPU, 4.8GHz, 32GB RAM, 10TB SSD), in order to have a dedicated server for the purposes of our research, and to have a resource for running all the conducted experiments and extracting results. By utilizing the node, we have unlimited access to all the transactions that took place since the first Ethereum block in July 2015, also called the Genesis block. We will refer to this dataset as the Node dataset. 
 
\begin{figure}[h]
    \centering
    \begin{subfigure}[b]{0.49\linewidth}
        \centering
        \includegraphics[width=\linewidth]{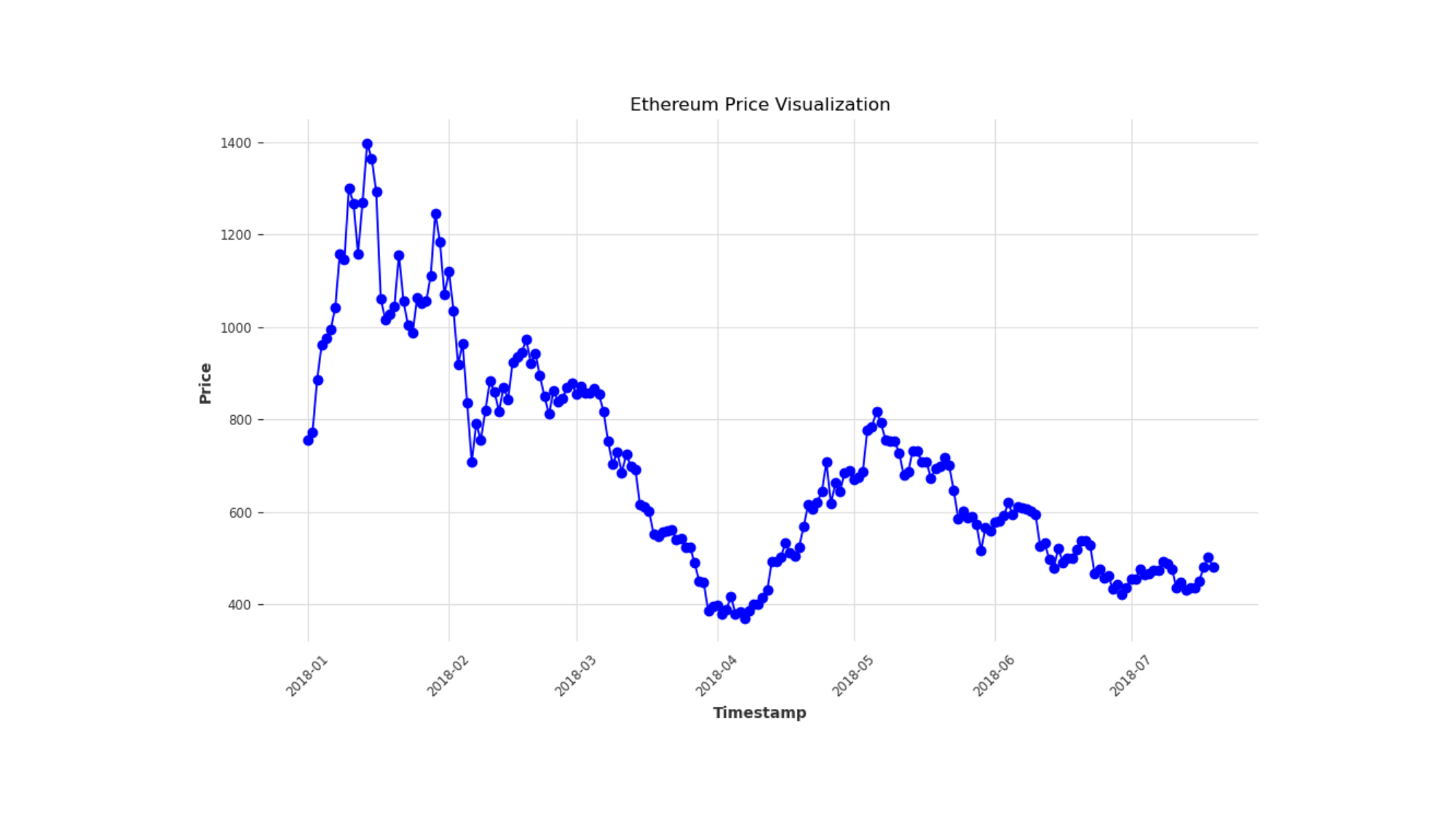}
        \caption{Kaggle Dataset}
        \label{fig:kaggle_dataset}
    \end{subfigure}
    \hfill
    \begin{subfigure}[b]{0.49\linewidth}
        \centering
        \includegraphics[width=\linewidth]{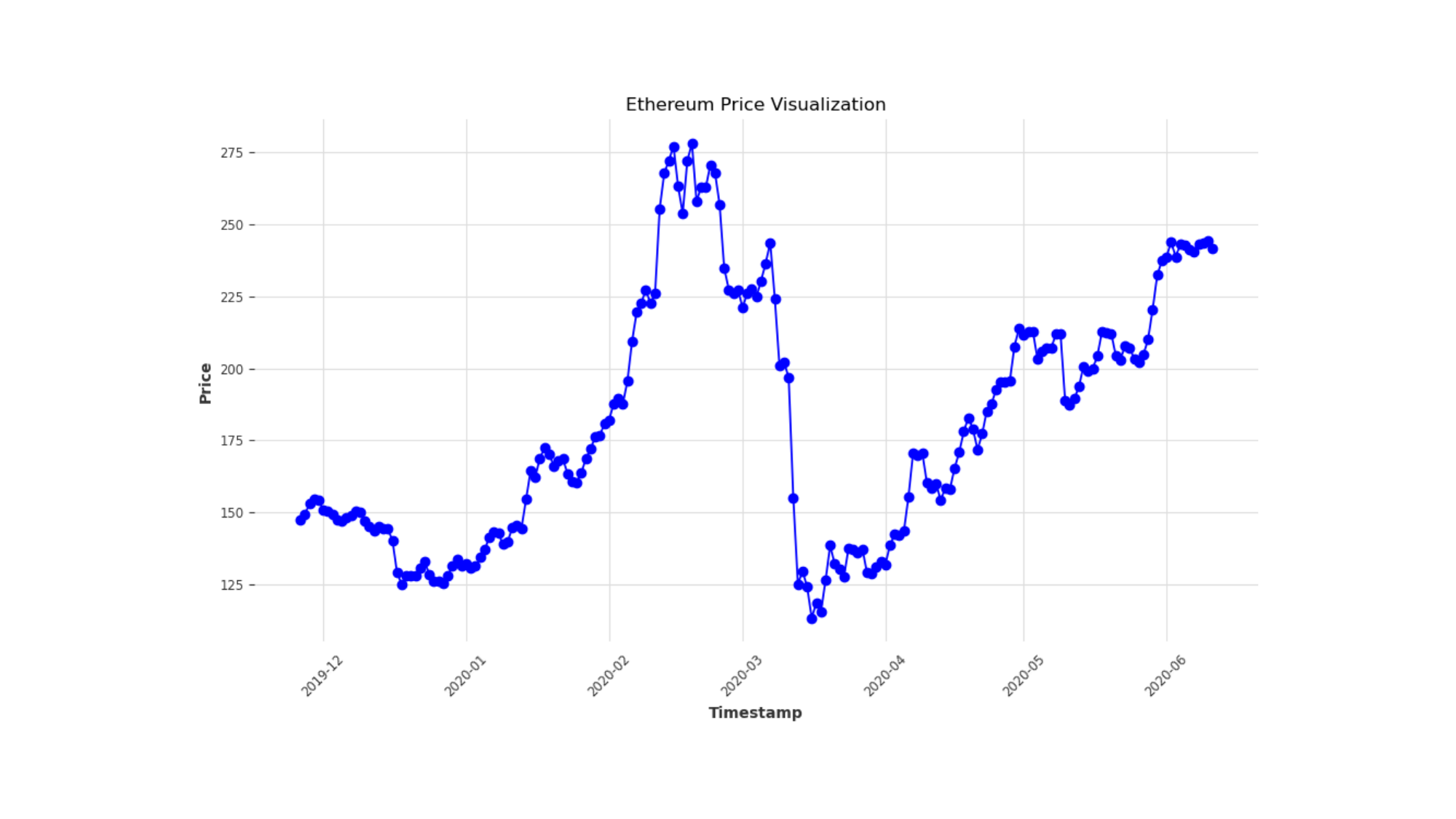}
        \caption{Node Dataset}
        \label{fig:proposed_dataset}
    \end{subfigure}
    \caption{Visualization of the volatility of Ethereum open price in different timestamps: Kaggle Dataset (a), Node Dataset (b).}
    \label{fig:dataset}
\end{figure}

Regarding the Ethereum price data, we used a publicly available Kaggle dataset \cite{ethereum_dataset_2023} which includes the following information: i) the daily price; ii) the opening price of ETH on the respective date (Dollars); iii) the highest price of ETH on the respective date (Dollars); iv) the lowest price of ETH on the respective date (Dollars); v) the Volume of ETH on the respective date (Dollars); vi) and the percentage of change from 2016 up to 2023. A visualization of the datasets is given in Figure \ref{fig:dataset}.

\subsection{Models and Fine Tuning}
This section provides an in-depth examination of the various models and fine-tuning approaches utilized in the current study as backbone models, focusing on their application to forecasting tasks. We explore three distinct modeling frameworks: Llama-3, Llama-2 and GPT-2 each chosen for its unique capabilities in handling timeseries data and predictive analytics. In order to apply the FPT logic, only LLMs with open source codes could be selected, which led us to the choice of GPT-2 and Llama. Furthermore, PatchTST was also employed in its initial form, as a baseline model, for further comparison.

\textbf{Llama-2}
 The Llama models, developed by Meta, are types of GPT models that are derived from the original Transformers architecture, while Llama-2 marks one of the newest innovation in open-source LLMs \cite{touvron2023llama}, \cite{roziere2023code}. These models use a pre-normalization method similar to GPT-3, incorporating the Root Mean Square (RMS) Normalization function at the beginning of each transformer sub-layer \cite{wu2024pmc}. Llama-2 also, uses a technique called grouped-query attention (GQA) to tackle memory bandwidth issues during the autoregressive decoding of Transformer models \cite{ainslie2023gqa}. This technique addresses the problem of needing to load decoder weights and attention keys/values at each processing step, which otherwise consumes a large amount of memory. GQA divides query heads (Q) into groups, each of which shares a single key head (K) and value head (V). This pre-normalization improves training stability by utilizing rescaling invariance and adapting the learning rate implicitly. Moreover, the LlamA-2 models replace the traditional ReLU activation function with the SwiGLU activation function, which enhances training efficiency and performance. 

 \textbf{Llama-3}
 Llama-3 \cite{meta2024llama3}, \cite{dubey2024llama} builds upon the foundation laid by LlaMA-2, offering significant advancements in performance and efficiency. One of the key enhancements in LlaMA-3 is the incorporation of multi-scale attention mechanisms, which allow the model to better capture temporal dependencies in time series data. Additionally, it introduces dynamic token routing \cite{kim-etal-2023-leap}, a technique that dynamically adjusts the attention allocation across input tokens, improving the model's ability to focus on the most relevant portions of the data. Unlike Llama-2, which employs GQA to address memory bandwidth issues, Llama-3 further optimizes memory efficiency by leveraging adaptive memory-sharing mechanisms, reducing computational overhead during autoregressive decoding. Another notable improvement is the use of an enhanced activation function, SwiGLU++, which further refines the training efficiency and stability of the model.

\textbf{GPT-2}
OpenAI introduced GPT, developed by Radford and Nara\-simhan \cite{radford2018improving}, which employs generative transformer decoders trained on extensive language corpora followed by fine-tuning for specific tasks. Subsequently, GPT-2, as detailed by Radford et al. \cite{radford2019language}, is trained on even larger datasets and with significantly more parameters, enhancing its adaptability to a variety of downstream tasks. Based on the framework of \cite{zhou2023one}, GPT-2 was fine-tuned for the forecasting task.

\textbf{PatchTST}
A patch-transformer-based model proposed in \cite{nie2022time}, namely PatchTST, was employed as a baseline model, in order to compare with the Llama and GPT-2 backbones FPTs. The PatchTST model, employs a design where each channel operates independently, with each one handling a distinct univariate time series. These series use a common embedding across all series. Additionally, the model features a patch design at the subseries level, which involves dividing the time series into smaller segments or patches. These patches then function as input tokens for the Transformer architecture.

\textbf{ANN, MLP, and LSTM}
ANNs \cite{zhou2017time}, MLPs \cite{gao2020application}, and LSTM \cite{jia2016investigation} networks are widely used machine learning models, each designed to address specific data modeling challenges. ANNs, composed of fully connected layers with non-linear activation functions, are effective for capturing non-linear relationships in data. MLPs extend ANNs by incorporating deeper architectures and dropout layers, enabling the learning of hierarchical patterns while mitigating overfitting. LSTMs, a type of recurrent neural network, excel in modeling temporal dependencies through their gating mechanisms, making them ideal for time series forecasting. These models serve as robust baselines in our experiments, in the following settings:
\begin{itemize}
    \item ANN: A two-layer fully connected architecture with 32 and 16 neurons, respectively, and ReLU activation functions, trained using the Adam optimizer with mean squared error as the loss function.
    \item MLP: A three-layer architecture with 64 and 32 neurons, incorporating dropout layers with a rate of 0.4 after each layer to prevent overfitting, and ReLU activations, trained using the Adam optimizer.
    \item LSTM: A single-layer LSTM with 50 units, followed by a dropout layer with a rate of 0.4, trained with the Adam optimizer and mean squared error loss to capture temporal dependencies effectively.
\end{itemize}

The architecture of the models, along with the framework pipeline which is analyzed in Section \ref{sec:pipeline}, are displayed in Figure \ref{fig:pipeline}.

\subsection{Model Pipeline}
\label{sec:pipeline}
\begin{figure*}[htbp]
\centering
\includegraphics[width=\linewidth]{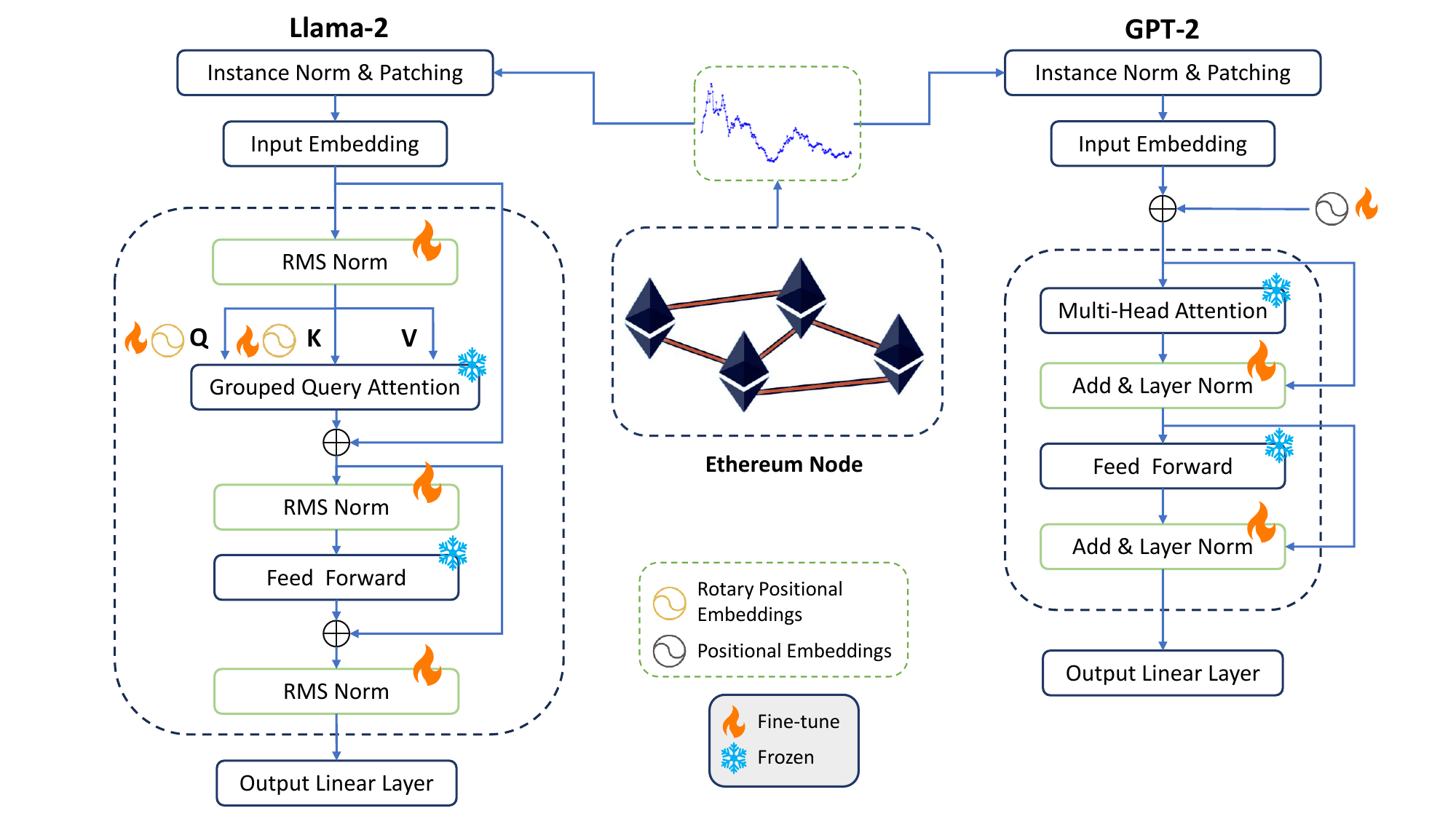}
\caption{Visualization of the models' architecture and pipeline. The architecture of our model pipeline incorporating Llama-2 (Left) and GPT-2 (Right) for Ethereum price prediction. The pipeline utilizes a combination of frozen and fine-tuned layers. For Llama-2, the grouped query attention and feed-forward layers are frozen, while rotary positional embeddings and RMS normalization layers are fine tuned. In GPT-2, the multi-head attention and feed-forward layers are frozen, while positional embeddings and layer normalization layers are fine-tuned. Data normalization and patching techniques are applied to enhance semantic information extraction. What applies to Llama-2, applies to Llama-3 as well.}

\label{fig:pipeline}
\end{figure*}

Our methodology adapts and extends the framework outlined in \cite{zhou2023one}, incorporating several key modifications to address our specific research objectives. Their implementation along with our alteration is given below:
\begin{itemize}
\item Model Selection: While Zhou et al. \cite{zhou2023one}, experimented only with GPT-2 as a backbone model, we expanded this work by utilizing the power of Llama as an Ethereum forecaster.
  \item Input Embedding: Zhou et al. \cite{zhou2023one} redesigned the embedding to fit time series data with linear probing \cite{alain2016understanding}, while the layer undergoes training to match pre-trained model dimensions. We use this technique, in order to project the time-series data to the required dimensions of both Llama and GPT-2.
    \item Frozen Pretrained Block (FPB): The main idea focuses on freezing self-attention layers and Feedforward Neural Networks (FFNs) to preserve learned knowledge, as these layers contain the majority of the knowledge encoded in pre-trained LLMs. 
    \item Positional Embeddings and Layer Normalization in GPT-2: As discussed in Sections \ref{sec:Intro} and \ref{related}, fine-tuning these components helps adapt the model to downstream tasks, particularly in our case, for Ether price prediction. 
    \item RMS Normalization layer and Rotary Embeddings in Llama: Inspired by the FPBs in GPT-2 model, we only fine-tune these elements in Llama-2 and Llama-3 implementation, to forecast the Ethereum prices.
    \item We implemented an extra custom-made class in order to handle data with daily basis intervals in the data loader function.
    \item Normalization and Patching: Based on their framework, a data normalization block was used, along with a patching technique \cite{nie2022time} to enhance semantic information extraction, allowing for efficient information processing and transfer. The normalization block, is performing a reverse instance norm \cite{kim2021reversible}, to enhance knowledge transfer. This normalization block normalizes the input time series using its mean and variance, and subsequently adds these values back to the output block.
\end{itemize}
The pipeline of the whole process, along with an example of Llama-2 and GPT-2 architecture, is visualized in Figure \ref{fig:pipeline}.
\subsection{Feature Handling}
In this paper, we primarily concentrate on univariate time series forecasting, exploring the capabilities and performance of LLMs in this domain within a short-term forecasting and few-shot setting.

Furthermore, incorporating multiple variables in a multivariate model increases the risk of overfitting \cite{faber2007avoid}, especially when the additional variables have low predictive power or when the dataset is limited. Univariate models, with their simpler structure, are less prone to this issue, promoting better generalization to unseen data.

The univariate approach simplifies the model's architecture and can lead to better performance metrics. By focusing on the primary feature—Ethereum price data—we were able to fine-tune the model more effectively. This adjustment proved crucial in achieving the state-of-the-art performance demonstrated in our results. In both Ethereum datasets, the targeted value was the "Open" price of the day.

\subsection{Forecasting Method}

\textbf{Short Term Forecasting} in cryptocurrency prediction is particularly valuable due to the highly volatile nature of cryptocurrency markets. The authors in \cite{alessandretti2018anticipating} demonstrate that short-term predictive models are crucial for Ethereum due to its volatile nature, which is often influenced by both broader crypto-market trends and Ethereum-specific developments. Cryptocurrency prices are highly sensitive to market sentiment, news, and external factors, which can cause significant price swings in a short time. Thus, short-term prediction is generally more effective than long-term prediction. Considering input length is crucial, as longer inputs are often believed to offer better results. However, in practice, some algorithms struggle to effectively utilize long input signals due to overfitting issues \cite{zhou2023one}. Thus, the sequence length for the input was set to 7 days, while the prediction length was set to one day. The dataset is split as 70\% for the training data, 20\% for the test data and 10\% for the validation data. To create input-output pairs from the dataset, we used a sliding window mechanism were we take the first 7 data points as the input sequence (historical "Open" prices) and we use the next prediction length points (in our case 1, because next day prediction) as the output sequence (future "Open" prices). Every time, we slide the window forward by 1 step and repeat until the end of the dataset.

\textbf{Few-Shot Forecasting}, a specialized area within meta-learning, is concerned with assessing tasks related to unfamiliar classes or datasets that only provide a very limited number of labeled instances \cite{wang2020generalizing}. The model learns from a vast set of related tasks, accumulating general knowledge applicable across various tasks. This foundational knowledge facilitates rapid adaptation to new, similar tasks. To thoroughly assess the representational capabilities of the Llama and GPT for Ethereum price prediction, we perform experiments within a few-shot learning framework. Likewise Short-term Forecasting, each time series is divided into three segments: training data, validation data, and test data but only a 10\% of the training data timesteps are utilized. In Few-shot Forecasting, training is halted after five epochs if there is no observed reduction in loss on the validation set, using an early stopping mechanism.

\subsection{Experimental Settings}
All experiments were conducted on a high-performance computing server equipped with four NVIDIA RTX A6000 GPUs, each with 48GB of VRAM. The environment utilized for the experiments included CUDA version 12.0.

\begin{table}[!t]
\renewcommand{\arraystretch}{1.3} % Adjusts the row height
\caption{Experimental Settings for Llama-3, Llama-2 and GPT-2}
\label{table_experimental}
\centering
\begin{tabular}{ccccl}
\toprule
\textbf{Settings} & \textbf{Llama-3} & \textbf{Llama-2} & \textbf{GPT-2} & \\ \midrule
Number of Layers & 80 & 80 & 12 \\
Hidden Size & 8192 & 8192 & 768 \\
Attention Heads & 64 & 64 & 12 \\
Max Embedding Position & 8192 & 4096 & 1024\\
Feed-Forward Dimension & 28672 & 28672 & 768 \\
Vocabulary Size & 128256 & 32000 & 50257 \\
Learning Rate & 1e-04 & 1e-04 & 1e-05 \\
Batch Size & 16 & 16 & 32 \\
Epochs & 20 & 20 & 20 \\
Optimizer & Adam & Adam & Adam \\
Patch size & 16 &16 &16 \\
Stride &8 &8 &8 \\
Activation function & SwiGLU++ &SwiGLU &GeLU \\
Early Stopping Patience & 5 & 5 & 5 \\

\bottomrule
\end{tabular}
\end{table}

Each model was trained using the Adam optimizer with the  batch size set to 16 for both Llama-3 70B and Llama-2 70B and 32 for GPT-2 for all experiments. We employed early stopping with a patience of 5 epochs based on the validation loss to prevent overfitting. Furthermore, the training loop uses mixed precision training with gradient accumulation, used to scale the loss for backpropagation and a learning rate scheduler that adjusts the learning rate using cosine annealing \cite{gotmare2018closer}. Specific details for the models' architecture as well as the hyperparameters used can be found in Table \ref{table_experimental}.

\section{Evaluation Criteria} \label{sec:evaluation}
The accuracy of models in price prediction tasks is typically evaluated using key performance indicators such as MSE, RMSE, and MAE. These metrics help in understanding the extent of error between the predicted and actual values, providing a comprehensive assessment of model performance \cite{patel2015predicting}. 
The formulas employed to calculate these statistical measures are detailed below:
\begin{align*} \text {MAE}=&\frac {1}{n}\sum _{k=1}^{n}{|y_{k} - \hat {y_{k}}|} \tag{1}\\ \text {MSE}=&\frac {1}{n}\sum _{k=1}^{n} (y_{k} - \hat {y_{k}})^{2} \tag{2}\\ \text{RMSE} = &\sqrt{\frac{1}{n}\sum_{k=1}^{n}(y_{k} - \hat{y_{k}})^{2}} \tag{3} \end{align*} 
where $y_k$ represents the actual values, and $\hat{y}_k$ denotes the forecasted ones.
\section{Results} \label{sec:results}
The results of GPT-2 and Llama showed SOTA performance in both Few-Shot and Short-term
Forecasting, as it is shown in Tables \ref{table_example} and \ref{table_example_FSF}.

In our analysis, we aimed to compare our results with three existing state-of-the-art works that utilized ANN, MLP, Transformers and LSTM models for Ethereum price prediction. However, two of those papers (\cite{kumar2020predicting}, \cite{kim2021predicting}) did not provide publicly available code implementations for their models. To ensure a fair and robust comparison, we replicated their methodologies as closely as possible based on the descriptions provided in their papers. Using the same dataset (Kaggle) and evaluation metrics, we implemented ANN, MLP, and LSTM models while following the reported architectural details, preprocessing steps, and hyperparameters wherever mentioned.

In the domain of Short-Term Forecasting, regarding the Kaggle dataset, our findings highlight the superior performance of Llama-3 as evidenced by the lowest MSE of 0.0027, indicating high predictive accuracy. The corresponding RMSE at 0.0518 further confirms the model's precision in predicting short-term fluctuations in Ethereum prices. GPT-2 achieved almost the same performance as Llama-3 and Llama-2 with an MSE score at 0.0029 and RMSE at 0.0524 for the same dataset. Furthermore, Llama-3 and Llama-2 outperformed GPT-2 in forecasting the Ethereum price of the Node dataset, with an MSE of 0.0042 and 0.0052 respectively, which, in comparison with Llama-3, is 0.0021 lower than GPT-2 and 0.0084 lower than PatchTST. These results significantly outperform the previously established benchmarks, such as the MLP and LSTM models by Kumar et. al \cite{kumar2020predicting} and the ANN from Kim et. al \cite{kim2021predicting}, as well as our own implementation of their models, which presented MSEs of 0.021, 0.018 and 0.0046, respectively. Additionally, PatchTST performed quite well with the Kaggle dataset but showed slightly worse results when predicting the "Open" price for the Node dataset. Interestingly, the application of Node Data with all the models, also showed promising results, but still not as remarkable as the Kaggle combination. This suggests that the choice of data source and especially the number of samples of each dataset, along with the employment of advanced language models, plays a critical role in enhancing forecasting accuracy in the volatile cryptocurrency market.

In the area of Few-Shot Forecasting, the results demonstrate the robust ability of the Llama and GPT-2 models to deliver accurate predictions with both Node Data and Kaggle Data, as reflected by the RMSEs. It is worth mentioning the impressive low MSE, which dropped at 0.030, when applied Llama-3 on the Kaggle dataset. These outcomes emphasize the models' capability at making reliable inferences from a minimal amount of data, which is of great importance in environments where swift analytical assessments are required due to limited information. As noted previously in the short-term forecasting task, Llama-3 performed better than GPT-2 with both datasets, but GPT-2 managed to outperform Llama-2 on the same experiments. However, it is worth noting that the integration of PatchTST with Node Data and Kaggle Data did not perform to the same standard, with significantly higher RMSEs of 0.7929 and 0.0823. This disparity in performance highlights the critical role that model selection plays in scenarios with sparse datasets, where Llama's and GPT-2’s advanced data processing capabilities markedly surpass the results of PatchTST, thereby establishing new leading standards for Few-Shot Forecasting within the cryptocurrency prediction landscape.

\begin{table}[!t]
\renewcommand{\arraystretch}{1.3} % Adjusts the row height
\caption{Short-Term Forecasting}
\label{table_example}
\centering
\begin{tabular}{cccccl}
\toprule
\textbf{Model} & \textbf{MSE} & \textbf{MAE} & \textbf{RMSE} & \textbf{Dataset}\\
\midrule
Transformer \cite{singh2024transformer} & 0.0062 & 0.0730 & 0.0791 & Kaggle\\

MLP \cite{kumar2020predicting} & 0.021 & 0.114 & 0.144  & \cite{kumar2020predicting} \\

LSTM \cite{kumar2020predicting} & 0.018 & 0.013 & 0.134 &  \cite{kumar2020predicting}\\

ANN \cite{kim2021predicting} & 0.0046 & - & 0.068 & \cite{kim2021predicting} \\

MLP  & 0.0122 & 0.0836 & 0.1100  & Kaggle \\

LSTM  & 0.0045 & 0.0450 & 0.0672 & Kaggle \\

ANN  & 0.0048 & 0.0495 & 0.0701 & Kaggle \\

%LSTM \cite{murray2023forecasting} & - & - & 0.0309 & 1825 \\

%Jagannath et al. \cite{jagannath2021chain}) & 0.6127 & 0.4755 & 0.7828 & 410 \\
%\hline
%Jay et al. \cite{jay2020stochastic}) & 0.6127 & 0.4755 & 0.7828 & 410 \\
%\hline
Llama-2 & 0.0044 & 0.0501 & 0.0696 & Node Data \\
Llama-3 & \textbf{0.0042} & 0.0494 & 0.0644 & Node Data \\

GPT-2 & 0.0063 & 0.0610 & 0.0796 & Node Data \\

PatchTST & 0.0126 & 0.0905 & 0.1167 & Node Data\\

Llama-2 & 0.0030 & 0.0385 & 0.0539 & Kaggle\\

Llama-3 & \textbf{0.0027} & 0.0355 & 0.0518 & Kaggle\\

PatchTST & 0.0052 & 0.0524 & 0.0733 & Kaggle\\
GPT-2 & 0.0029 & 0.0373 & 0.0524 & Kaggle\\

\bottomrule
\end{tabular}
\end{table}

\begin{figure}[h!]
    \centering
    \begin{subfigure}[t]{0.49\linewidth}
        \centering
        \includegraphics[width=\linewidth]{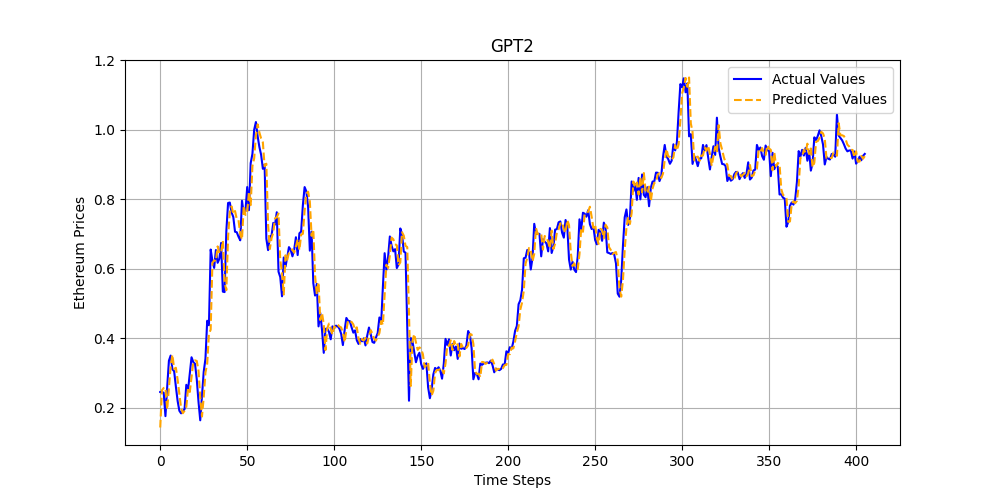}
        \caption{GPT2 - Short-term Forecasting}
        \label{fig:long_gpt2}
    \end{subfigure}
    \hfill
    \begin{subfigure}[t]{0.49\linewidth}
        \centering
        \includegraphics[width=\linewidth]{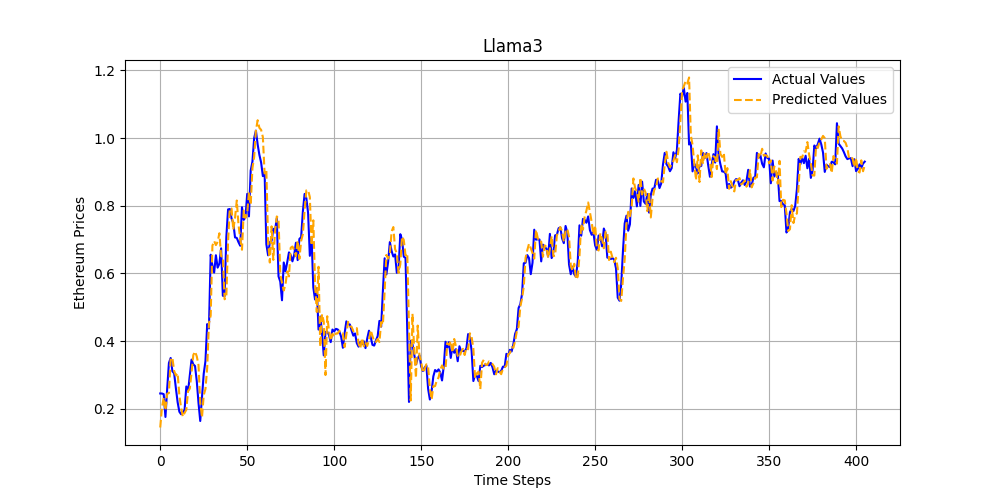}
        \caption{Llama3 - Short-term Forecasting}
        \label{fig:long_llama3}
    \end{subfigure}

    \vspace{0.5cm} % Space between rows

    \begin{subfigure}[t]{0.49\linewidth}
        \centering
        \includegraphics[width=\linewidth]{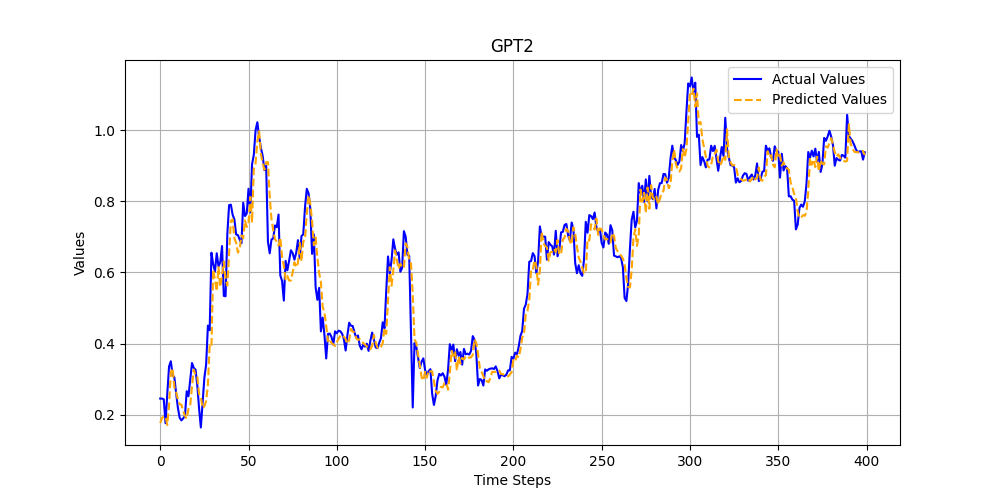}
        \caption{GPT2 - Few-Shot Forecasting}
        \label{fig:few_gpt2}
    \end{subfigure}
    \hfill
    \begin{subfigure}[t]{0.49\linewidth}
        \centering
        \includegraphics[width=\linewidth]{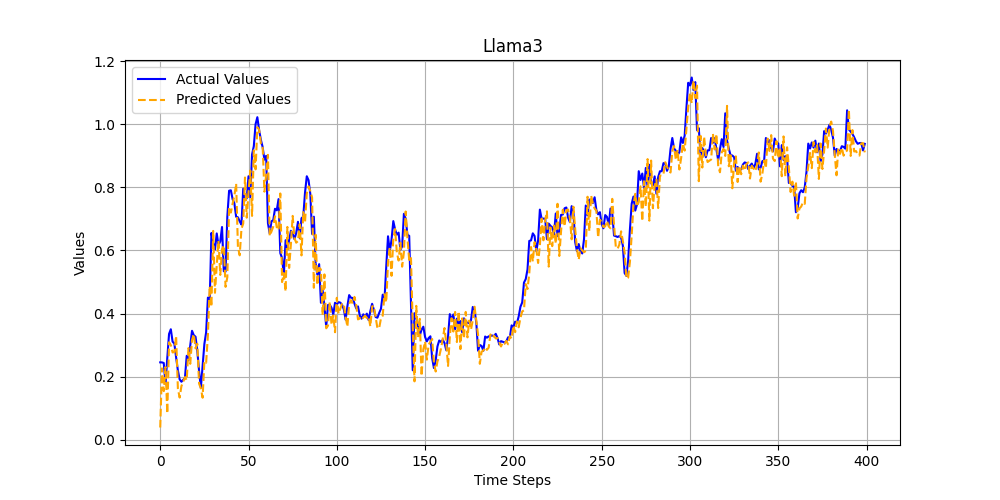}
        \caption{Llama3 - Few-Shot Forecasting}
        \label{fig:few_llama3}
    \end{subfigure}

    \caption{Visualization of Ethereum price predictions depicting the actual and predicted values in both short-term and few-shot forecasting for the Kaggle dataset: (a) GPT2 in short-term forecasting and (b) Llama-3 in short-term forecasting c) GPT2 in Few-shot forecasting and d) Llama-3 in Few-shot forecasting.}
    \label{fig:comparison}
\end{figure}

\begin{table}[!t]
\renewcommand{\arraystretch}{1.3} % Adjusts the row height
\caption{Few-Shot Forecasting}
\label{table_example_FSF}
\centering
\begin{tabular}{ccccl}
\toprule
\textbf{Model} & \textbf{MSE} & \textbf{MAE} & \textbf{RMSE} \\ \midrule
Node Data \& Llama-2 & 0.0075 & 0.0806 & 0.0915 \\
Node Data \& Llama-3 & \textbf{0.0065} & 0.0609 & 0.0885 \\
Node Data \& GPT2 & 0.0074 & 0.0638 & 0.0910 \\
Node Data \& PatchTST & 0.3519 & 0.3987 & 0.7929 \\
Kaggle Data \& Llama-2 & 0.0041 & 0.0489 & 0.0686 \\
Kaggle Data \& Llama-3 & \textbf{0.0030} & 0.0379 & 0.0566 \\
Kaggle Data \& GPT-2 & 0.0036 & 0.0435 & 0.0652 \\
Kaggle Data \& PatchTST & 0.0066 & 0.0591 & 0.0823 \\

\bottomrule
\end{tabular}
\end{table}
To complement the numerical evaluation of our models, we provide visualizations that compare predicted and actual Ethereum prices in Figure \ref{fig:comparison}. Such visual comparisons are crucial for illustrating the models’ ability to adapt to the highly volatile nature of cryptocurrency prices. As highlighted in \cite{alessandretti2018anticipating}, the dynamic and non-linear behavior of cryptocurrency markets makes it essential to evaluate forecasting models not only through statistical metrics but also by examining their behavior in visualized scenarios. This approach provides clearer insights into the model's performance, highlighting strengths and limitations that may not be evident from metrics alone. By including these visualizations, we aim to offer a more comprehensive evaluation of the forecasting capabilities the models under volatile conditions.

\section{Conclusion} \label{sec:conclusion}
In this research, we have explored the potential of LLMs, specifically Llama models and GPT-2, for predicting Ethereum prices, demonstrating their effectiveness in both short-term and few-shot forecasting scenarios. Our study has adapted and extended the framework in \cite{zhou2023one}, based on the method of freezing the pre-trained transformer layers, to keep the majority of the knowledge, and fine-tuning on the rest, to better suit the intricacies of cryptocurrency market data.

We have demonstrated through various experiments that LLMs, when properly fine-tuned and adapted to the domain-specific requirements of financial time series data, can indeed compete with or even outperform traditional state-of-the-art models. This is especially evident in our results, where Llama-3 showcased superior performance, achieving lower error rates across key metrics like MSE, MAE, and RMSE compared to other evaluated models.

Moreover, it would be interesting to investigate both on-chain and off-chain factors, with a particular focus on incorporating sentiment data from social media platforms to enhance predictive accuracy, including those from Reddit, Twitter, and Google Trends, among other sources. We plan to expand our work by enriching the datasets with extra features and by experimenting with different state-of-the-art LLMs and frameworks, with focus in the timeseries prediction task.

\section{Acknowledgments}
This paper was prepared in part for information purposes by the AI Research Group, the AlgoCRYPT Center of Excellence, and Cybersecurity \& Technology Controls group of JPMorgan Chase \& Co and its affiliates (JP Morgan), and is not a product of the Research Department of JP Morgan. JP Morgan makes no representation and warranty whatsoever and disclaims all liability, for the 8 completeness, accuracy or reliability of the information contained herein. This document is not intended as investment research or investment advice, or a recommendation, offer or solicitation for the purchase or sale of any security, financial instrument, financial product or service, or to be used in any way for evaluating the merits of participating in any transaction, and shall not constitute a solicitation under any jurisdiction or to any person, if such solicitation under such jurisdiction or to such person would be unlawful.

\bibliographystyle{unsrt}  
\bibliography{references}

\end{document}